%% file: main.tex
\documentclass[11pt]{article}

\usepackage[T1]{fontenc}

\usepackage{helvet}
\usepackage[margin=1in]{geometry}
\usepackage{titlesec}
\usepackage{graphicx}
\usepackage{parskip}
\usepackage[dvipsnames]{xcolor}
\usepackage{tcolorbox}
\usepackage[numbers,square]{natbib}  
\usepackage{amsmath, amssymb}
\usepackage{fancyhdr}
\usepackage{microtype}
\usepackage{enumitem}
\usepackage{hyperref}  

\usepackage{url}
\usepackage{booktabs}
\usepackage{amsfonts}
\usepackage{nicefrac}
\usepackage{float}
\usepackage{algorithm}
\usepackage{algpseudocode}
\usepackage{inconsolata}
\usepackage[labelfont=bf]{caption}
\usepackage{multirow}
\usepackage{refcount}
\usepackage{footmisc}
\usepackage{wrapfig}

\usepackage{tcolorbox}

\definecolor{metaBlue}{RGB}{24,119,242}
\definecolor{abstractBoxBG}{RGB}{245,245,250}

\fancypagestyle{firstpagestyle}{
  \fancyhf{}
  \rhead{\textsf{\small Stanford NeuroAI Lab}}
  \cfoot{\thepage}
}
\titleformat{\section}{\color{black}\sffamily\Large\bfseries}{}{0pt}{}
\titlespacing*{\section}{0pt}{1.5\baselineskip}{1\baselineskip}

\titleformat{\subsection}{\color{black}\sffamily\large\bfseries}{}{0pt}{}
\titlespacing*{\subsection}{0pt}{1.5\baselineskip}{1\baselineskip}

\usepackage{xspace}

\begin{document}

\vspace*{-2em}
\begin{tcolorbox}[
  colback=abstractBoxBG,
  colframe=abstractBoxBG,  
  boxrule=0pt,
  arc=4pt,
  left=12pt, right=12pt, top=10pt, bottom=12pt,
  width=\textwidth,
  enlarge left by=0mm,
  enlarge right by=0mm
]

{\sffamily\LARGE\bfseries Model-brain comparison using inter-animal transforms \\[0.5em]}
{\sffamily\textbf{Imran Thobani}$^{*,1}$, \textbf{Javier Sagastuy-Brena}$^{2}$, \textbf{Aran Nayebi}$^{3}$, 
\textbf{Jacob Prince}$^{4}$,\\
\textbf{Rosa Cao}$^{1}$, \textbf{Daniel Yamins}$^{*,1}$} \\[0.5em]
\small
$^1$Stanford University, 
$^2$Opal Camera,
$^3$Carnegie Mellon University,
$^4$Harvard University\\[1em]
\textbf{Abstract.} Artificial neural network models have emerged as promising mechanistic models of the brain. However, there is little consensus on the correct methods for comparing activation patterns in these models to brain responses.
Drawing on recent work on mechanistic models in philosophy of neuroscience, we propose a comparison methodology based on the Inter-Animal Transform Class (IATC) - the \emph{strictest set of functions needed to accurately map neural responses between real subjects (or ``animals'') in an empirical population}.
Using the IATC, we can map bidirectionally between a candidate model's responses and brain data, assessing how well the model can masquerade as a typical subject using the same kinds of transforms needed to map across real animal subjects. 
We empirically identify the IATC in three settings: a simulated population of neural network models, a population of mouse subjects, and a population of human subjects. 
We find that the IATC resolves detailed aspects of the mechanism of a neural computation, such as the non-linear activation functions present in the visual cortical hierarchy. 
Most importantly, we find that the empirical IATC enables highly accurate predictions of neural activity while also achieving high specificity in mechanism identification, evidenced by its ability to \emph{separate} response patterns from different brain areas while strongly \emph{aligning} same-brain-area responses between subjects.
In other words, the IATC is a proof-by-existence that there is no inherent tradeoff between the neural engineering goal of high model-brain predictivity and the neuroscientific goal of identifying mechanistically accurate brain models.   
Using IATC-guided transforms, we obtain new evidence, convergent with previous findings, in favor of topographical deep neural networks (TDANNs) as models of the visual system.
Overall, the IATC enables principled model-brain comparisons, contextualizing previous findings about the predictive success of deep learning models of the brain, while improving upon previous approaches to model-brain comparison.
\\

\textsuperscript{$*$}Corresponding authors: ithobani@stanford.edu, dyamins@gmail.com.

\end{tcolorbox}

\vspace{2em}

\vspace{-2.5em}

\input{sections/intro}

\input{sections/iatc}

\input{sections/results}

\input{sections/discussion}

\clearpage

\section*{Acknowledgements}

This work was supported by the following awards: To I.T.: Patrick Suppes Philosophy of Science Dissertation Award. To D.L.K.Y.: Simons Foundation grant 543061, National Science Foundation CAREER grant 1844724, National Science Foundation Grant NCS-FR 2123963, Office of Naval Research grant S5122, ONR MURI 00010802, ONR MURI S5847, and ONR MURI 1141386 - 493027. We also thank the Stanford HAI, Stanford Data Sciences and the Marlowe team, and the Google TPU Research Cloud team for computing support. 

\input{sections/suppl}

\bibliographystyle{unsrtnat}
\bibliography{references}

\end{document}

%% file: sections/intro.tex
\section{Introduction}
Artificial neural network (ANN) models have been found to exhibit internal activation patterns that predict aspects of neural activity in a wide variety of brain areas~\citep{zipser1988back, olshausen1996emergence, yamins2014performance, storrs2021diverse, kell2018task, zhuang2021unsupervised, khaligh2014deep, Sussillo2015, wang2021evolving,schrimpfFedorenkoLanguage,mineaultDorsal,nayebi2021MEC, nayebiPhysicalReasoning}.
These results naturally raise the question of whether these models can serve as mechanistic models of brain function, at least at some level of abstraction \citep{cao2021explanatory, kriegeskorte2016inferring}.
Although a variety of methods have been proposed for quantitatively assessing neural response similarity between models and brains \citep{sucholutsky2023getting}, there is little consensus on what the correct method is.

\begin{figure}
    \includegraphics[width=\linewidth]{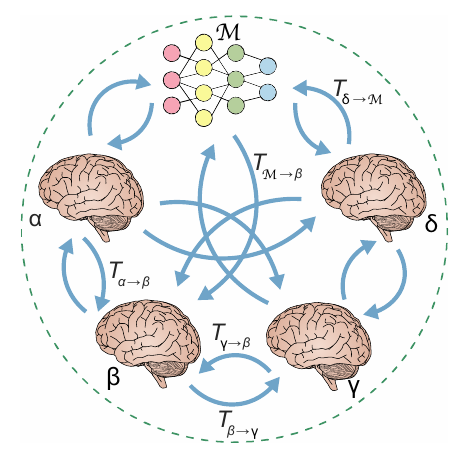}
    \caption{\textbf{Model-brain comparison using inter-animal transforms.}
    The Inter-Animal Transform Class (IATC) is the strictest set of functions required to map responses accurately between subjects in a population for a given brain area. 
    We propose to use the empirically identified IATC to map bidirectionally between a candidate model's responses and brain responses in order to assess whether the model can masquerade as a typical animal subject.
    The arrows represent individual functions in the IATC.
    These functions may not in general be invertible, so the arrows are not symmetric.
    Nonetheless, bidirectional mapping emerges from the fact that it is possible to map responses between any pair of subjects in the natural population using the IATC, and thus we map in both directions when comparing a model to the brain, just as we would when comparing two actual brains. 
    This allows us to separate between models that do match the brain under the IATC and those that do not.
    }
    \label{fig:iatc_schematic}
\end{figure}

Because a good mechanistic model of the brain should accurately predict the neural activities (e.g. firing rates) of typical subjects, we focus our attention on predictive mappings that relate unit activations in the model to neuronal firing rates, as opposed to summary statistics methods, like Representational Similarity Analysis, that compare second-order statistics of neural responses \cite{kriegeskorte2008representational, harvey2023duality}.
Mapping-based approaches assess model-brain similarity in terms of how well model responses predict brain responses (and vice versa) under a chosen class of functions used to map activation patterns between sets of neurons or units.

Although it is an informal assessment, mapping methods are often judged by the ``strictness" or ``flexibility" of the class of transforms they employ.
A particularly stringent approach to mapping models to brains would be to use 1-1 matching, i.e. attempting to bijectively identify each ANN model unit with a unique neuron in a target animal's brain.
However, a key challenge complicating model-brain mapping is the fact that the target of modeling is not a single idealized brain, but rather a \emph{population} of brains that are all somewhat different from each other.
In fact, inter-subject variability can be substantial -- in humans, the estimated number of neo-cortical neurons can vary between subjects by up to a factor of 2 \citep{haug1987brain}, and even the exact number of functionally identified brain areas can vary \citep{gao2022genuine}.
As a result, 1-1 matching is likely to be problematic when mapping a single model to different subjects in a population.
A more sophisticated approach to model-brain mapping is therefore needed.

A first generation of approaches to this problem used linear mappings to compare models to brains~\citep{yamins2014performance}.
By introducing some flexibility into the mapping class, linear regression enables accurate predictions when mapping a given model to typical subjects.
However, concerns have arisen that the linear mapping class, precisely because it is relatively flexible, is worse for specificity, i.e. the ability of the mapping class to distinguish the target neural mechanism from alternative mechanisms ~\citep{kornblith2019similarity, ding2021grounding, conwell2022}.  
As a result, recent discussions have proposed stricter mapping classes, such as soft matching~\citep{khosla2023soft}, which matches individual neurons between neural populations as 1-1 matching does, while also allowing for neural populations of different sizes.
Because of their sensitivity to individual neuron activities, these stricter methods have been thought to be more effective at distinguishing the target mechanism from alternatives (i.e. specificity), even if less effective at predicting neural responses for different subjects. 
However, given that predictive accuracy and specificity are both critical for model-brain comparison, it has remained unclear what the right mapping class is.

Here we develop the idea of the \emph{\textbf{Inter-Animal Transform Class}} (IATC), a concept that has been introduced in philosophy of neuroscience to handle the problem of between-subject variability when building mechanistic models~\citep{cao2021explanatory}.
Informally, the IATC is the  set of mappings needed to align brains to another within a population (e.g. a population of typical mice).
The core principle of our approach is to \textbf{align models to brains using the very same mappings needed to align brains to each other} - in other words, using the empirically-identified IATC (Fig.~\ref{fig:iatc_schematic}).

%% file: sections/iatc.tex
\section{The Inter-Animal Transform Class}
In this section, we provide a formal definition of the IATC, which in turn enables concrete metrics for estimating the IATC for a given population.

\begin{tcolorbox}[colback=white, colframe=black, boxrule=1pt, title=Definition: Inter-Animal Transform Class (IATC)]
    The IATC is defined as the strictest (smallest) set of functions that maps neural responses between subjects in a natural population with as high accuracy as possible.
\end{tcolorbox}

Both mapping accuracy and strictness are important criteria for the IATC.
First, the IATC must achieve the maximum possible mapping accuracy across subjects in the population.\footnote{The maximum possible accuracy that is obtainable may be less than perfect because different subjects' neural representations can have different metamers \citep{feather2023model}, and therefore different neural encoding functions with different null spaces.} 
Second, subject to this mapping constraint, the IATC must be as strict as possible, meaning it contains \textit{only} the functions that are necessary to achieve high mapping accuracy, thus ensuring that the equivalence relation between neural response profiles defined by the IATC is as strong as possible.
More formally, a set of functions $A$ is defined as  \textit{stricter} than a set of functions $B$ if $A$ is a proper subset of $B$.
As an example, the set of 1-1 mappings (i.e. permutations) is stricter than the class of linear mappings, since any 1-1 mapping is also a linear mapping, but the reverse is not true.
More generally, stricter mapping classes impose tighter constraints, such as requiring small groups of neurons to be matched across subjects.

Though defined relative to a population of real individuals, the IATC can be used to compare artificial networks to brains.  
Specifically, we propose to use the empirically-identified IATC itself to map between models and brains -- in effect, measuring how well the model can masquerade as a member of the population (Fig. \ref{fig:iatc_schematic}).  
A key implication of this IATC-based approach is that model-brain mappings should be performed \emph{bidirectionally} between models and brain data, just as when comparing two brains to each other, rather than only mapping in one direction (from model to brain). 

\begin{figure}[h!]
    \begin{center}
    \includegraphics[width=\columnwidth]{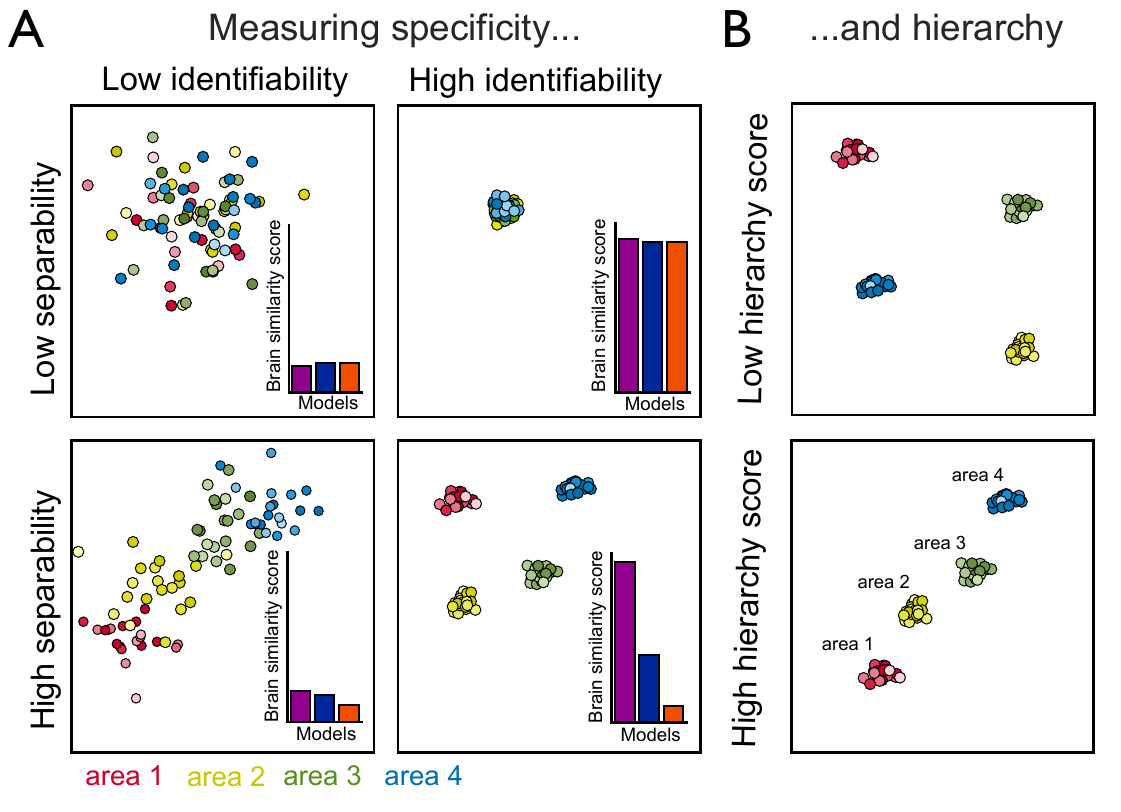}
    \end{center}
    \caption{\textbf{Evaluating candidate transform classes for specificity and hierarchy.}
    (A) It is desirable for a mapping to achieve high \emph{specificity} -- simultaneously achieving within-area \emph{identifiability} and between-area \emph{separability} when mapping responses between animals.
    Each dot is a response profile for a particular subject and brain area.
    The schematic barplots represent likely outcomes for model separation when comparing models to brains.
    (B) To capture more graded relationships beyond specificity, we also look at the correlation between dissimilarity scores and distances in a known hierarchy.
    }
    \label{fig:specificity}
\end{figure}

With the formal definition of the IATC in hand, the primary challenge is how to actually estimate the IATC for a given population of subjects. 
Estimating the IATC for a given population requires jointly optimizing both mapping accuracy and strictness over the space of all possible transform classes.
In cases where the correct brain parcellation is known in advance, the search space of candidate IATCs is constrained by the fact that the true IATC must map responses across subjects using only neurons from the \textit{same} brain area (by virtue of strictness) without losing mapping accuracy.\footnote{In the Discussion section, we return to the issue of what to do in cases where the correct brain parcellation is unknown. As we argue there, IATC estimation does not inherently require knowing the correct brain parcellation in advance, and in fact can potentially refine and discover brain parcellations.}
Therefore, a good IATC candidate must succeed in \emph{aligning} same-area responses across subjects, while being strict enough to \emph{separate} responses from different areas. 
This pair of desiderata naturally suggests a meta-metric of \textit{specificity}, which evaluates the extent to which a mapping simultaneously achieves high cross-subject within-area identifiability while maintaining between-area separability (Fig. \ref{fig:specificity}A).
As an extension of specificity, we also consider whether similarity scores under a mapping method correlate with inter-area distances in a known hierarchy (Fig. \ref{fig:specificity}B).

Ideally, we would estimate the IATC directly using large-scale optimization techniques applied to a massive neural dataset with many subjects.
In the absence of the required data and techniques for doing so, here we instead evaluate a spectrum of well-known methods, such as linear regression and soft matching, against our two metrics for the quality of an IATC estimate: they should map responses across subjects within an area as accurately as possible, while separating responses from different areas.

We first evaluate candidate transform classes on a simulated population of artificial neural network models. 
Because we have complete knowledge of the network structure and can ``measure'' responses for all units over many stimuli, we are able to observe how the specific form of the activation function in the network shapes the relationships between model subjects' responses.
This motivates a new transform class that maps responses across model subjects with close-to-maximum predictivity and high specificity, yielding a reasonable estimate of the IATC.
We then evaluate these different methods on real neural datasets, including both mouse electrophysiology and human fMRI recordings.

%% file: sections/results.tex
\section{Results}
\label{sec:results}

\subsection*{\center{Testing candidate IATCs for a simulated population}}
\label{ssec:simulated_population}
We first evaluate transform classes on a simulated population of neural networks (Figure \ref{fig:conceptual-pre-post}A) against the IATC criteria by testing for same-area predictivity as well as for specificity.
Our simulated population consists of neural networks based on a state-of-the-art model of mouse visual cortex: an AlexNet trained with contrastive learning on 64x64 inputs \citep{nayebi2022mouse}.
We further modified the model to use a softplus activation function followed by Poisson-like noise to better mimic neuronal response characteristics. 
To generate a population of model subjects, we vary the random seed controlling the weight initialization and training data order.
We map responses between subjects using 10000 activation patterns driven by ImageNet-validation stimuli (80/20 train-test split), evaluating the test $R^2$, median across target neurons for a given model layer, averaged in both directions and across all pairs of subjects.

\subsubsection*{\textit{The activation function has a substantial effect on same-layer response similarity between model subjects.}}
\begin{figure*}[h!]
    \begin{center}
    \includegraphics[width=\linewidth]{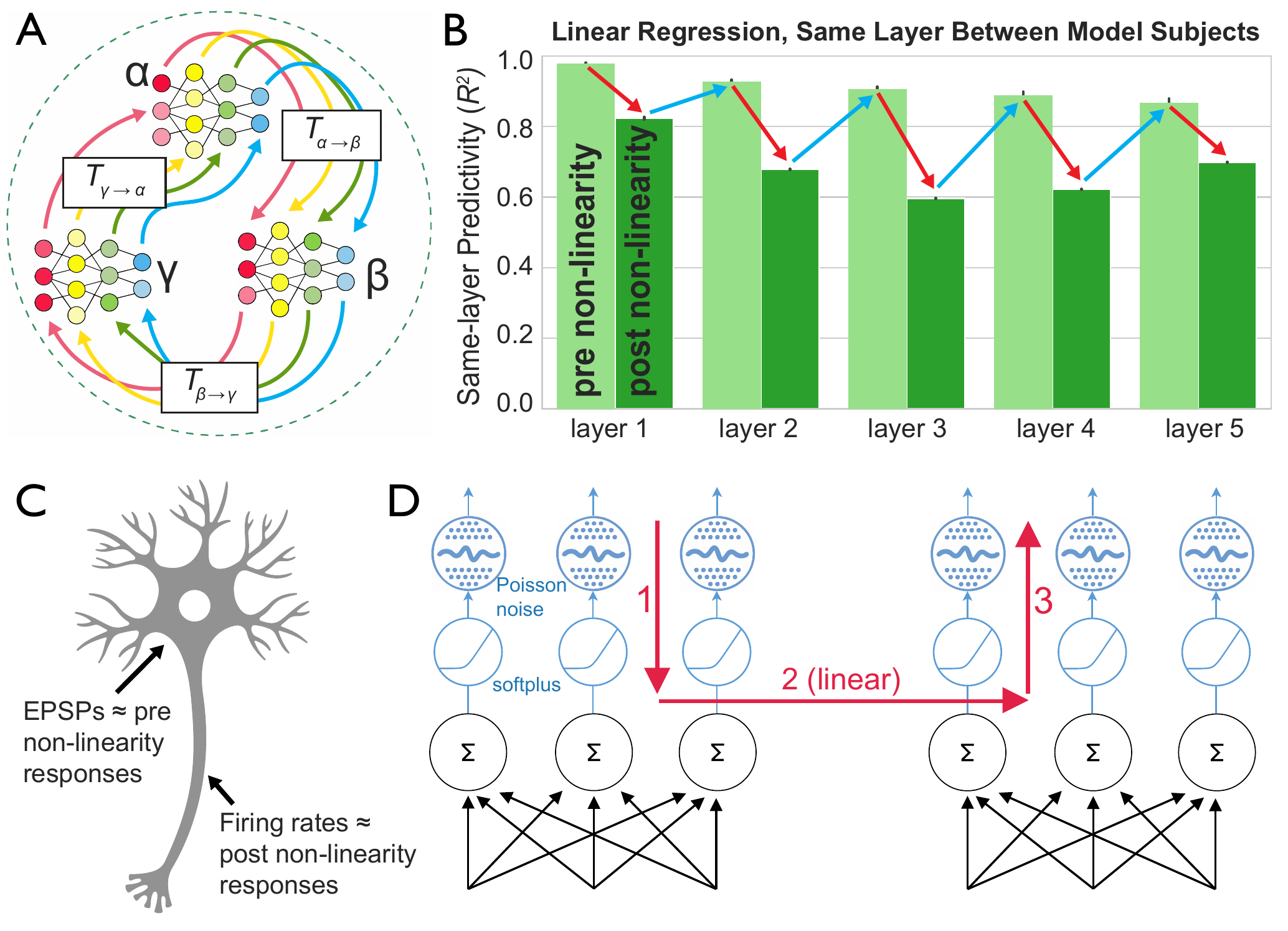}
    \end{center}
    \caption{\textbf{Assessing same-area similarity in the model population reveals a ``zippering'' effect caused by the model activation function.}
    (A) We attempt to identify the IATC for a model population by first assessing within-area similarity when mapping between differently seeded model subjects. 
    (B) \textbf{The Zippering Effect:} Post-non-linearity activations are only moderately similar between subjects up to linear transform, especially for the intermediate layers. 
    However, pre-non-linearity activations are highly similar at all layers. 
    At every layer, the non-linear activation function causes responses between different subjects to diverge somewhat up to linear transform, only for responses to converge again at the next layer's pre-activations, before diverging and converging again.
    (C) Post-non-linearity responses can be thought of as corresponding to firing rates, while pre-non-linearity responses can be thought of as corresponding to EPSPs (excitatory post synaptic potentials).
    (D) \textbf{The Zippering Transform:} The Zippering Effect provides a clue as to how to construct a better transform class - namely, by accounting for details of the mechanism (the non-linear activation function).
    Step 1 inverts the non-linearity to recover the pre-non-linearity activations of one subject, step 2 applies a fitted linear transform to predict the pre-non-linearity activations of the other subject, and step 3 re-applies the non-linearity to predict post-non-linearity activations.
    }
    \label{fig:conceptual-pre-post}
\end{figure*}

Because a viable IATC candidate must map responses accurately across different subjects for the same layer, we first evaluate linear regression, which has been widely used for neural response prediction \citep{canatar2024spectral}.
Specifically, we use cross-validated ridge regression.
Surprisingly, linear regression achieves only moderate same-layer predictivity for intermediate model layers when mapping post-softplus activations between subjects (Fig. \ref{fig:conceptual-pre-post}B).
This raises the possibility that the IATC might require highly non-linear transforms such as those implemented by an MLP, which if true would suggest that model subjects trained from different random seeds are highly dissimilar in their learned representations.

However, we observe a ``zippering'' effect: at each layer, pre-non-linearity responses are close to linearly related between subjects, but the activation function disrupts these linear relationships for post-non-linearity responses, before the next layer's pre-non-linearity responses re-converge up to a linear transform (Figure \ref{fig:conceptual-pre-post}B).
This effect suggests that the model subjects are actually similar, despite the apparent divergence suggested by the failure of linear regression to map post-non-linearity responses accurately.
Pre-non-linearity activations can be thought of as corresponding to trial-averaged EPSPs in real neurons, while post-non-linearity activations can be thought of as corresponding to trial-averaged firing rates (Figure \ref{fig:conceptual-pre-post}C). 
Because EPSPs are hard to measure, we develop an IATC candidate that works for post-non-linearity responses.

\subsubsection*{\textit{Improving cross-subject mapping by considering the non-linear activation function.}}
\label{ssec:mechanism}

\begin{figure*}[htbp]
    \begin{center}
    \includegraphics[width=\linewidth]{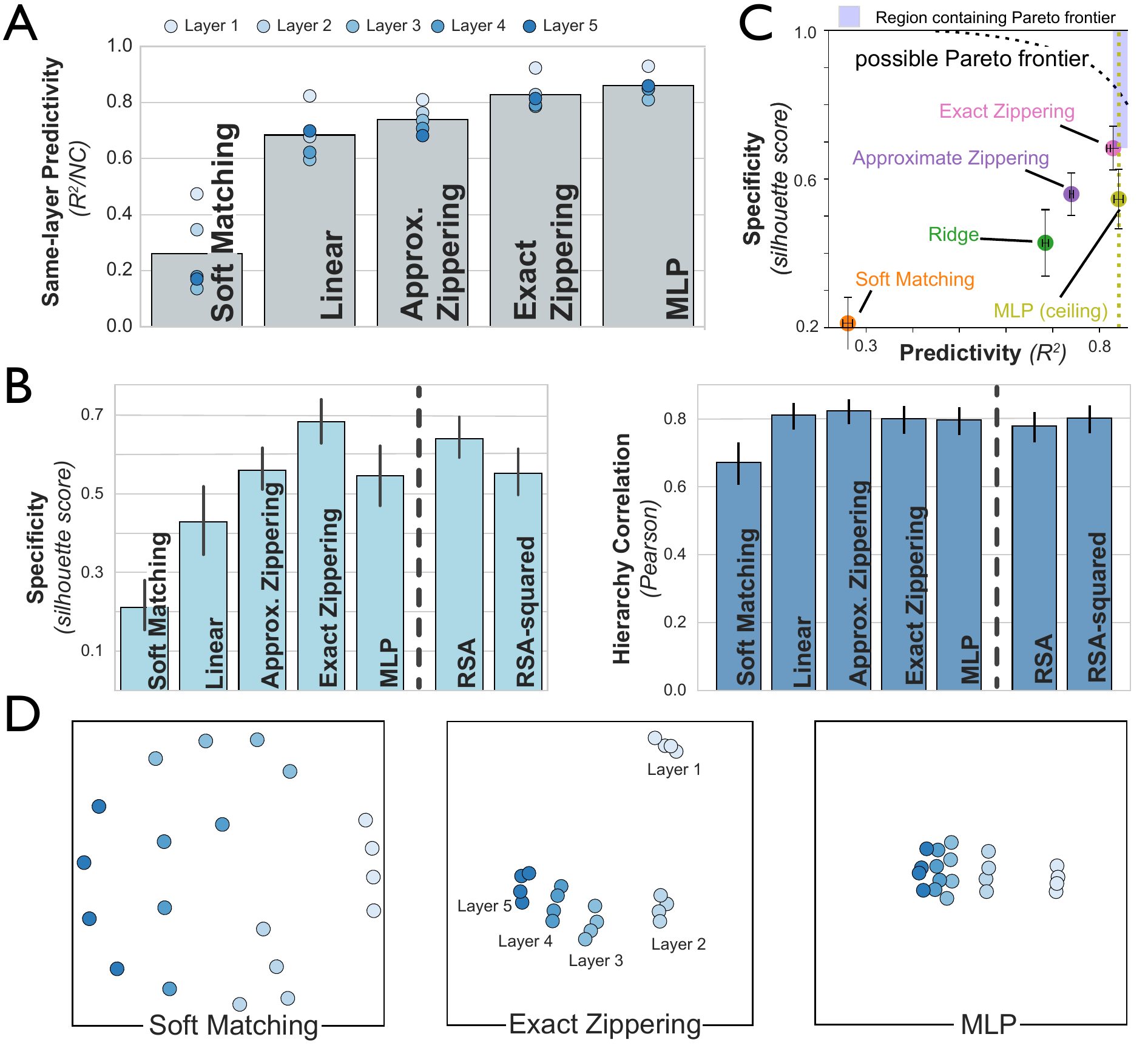}        
    \end{center}
    \caption{\textbf{Predictivity and specificity for a spectrum of candidate transform classes on a simulated model population.
    }
    (A) Same-layer predictivity when mapping responses between model subjects.
    (B) Specificity and hierarchy correlation for different transform classes.
    Error bars are 95\% CIs.
    (C) A scatterplot comparing predictivity and specificity across transform classes.
    While the exact shape of the Pareto frontier for predictivity and specificity is unknown, we identify a bounded region (shaded blue) that contains at least one Pareto-optimal point. 
    The diagonal of this region represents the maximum possible distance from our best IATC candidate (Exact Zippering) to the Pareto frontier.
    (D) Multidimensional scaling (MDS) plots to visualize dissimilarity scores, when mapping response profiles between all layers and all subjects.
    Each dot is a response profile for a particular subject and model layer. 
    Distances between dots are optimized to match the dissimilarities using a particular comparison method.
    }
    \label{fig:model_results}
\end{figure*}

The zippering effect suggests that an ideal IATC candidate must consider the effect of the non-linearity on the relationships between different subjects' post-non-linearity responses.
We develop the Zippering Transform, which inverts the softplus activation function, applies a fitted linear mapping between the two subjects, and re-applies the softplus activation to predict the target subject's post-softplus responses (Fig. \ref{fig:conceptual-pre-post}D).
Building on an established framework of generalized linear models (GLMs) \citep{mccullagh2019generalized, chichilnisky2001simple}, we use a GLM whose inverse link function is precisely matched to the softplus activation in order to fit the linear mapping and re-apply the softplus activation function.
Exactly accounting for the Zippering Effect in this way yields substantially higher predictivity than linear regression when mapping post-softplus activations (Fig. \ref{fig:model_results}A, Exact Zippering).

In the case of real neural data, we may not know the exact form of the activation function, and we therefore develop a version of the Zippering Transform that attempts to approximately account for the activation function.
Approximate Zippering approximately inverts the activation function (step 1 of Fig.~\ref{fig:conceptual-pre-post}D) for the source model subject by using Yeo-Johnson scaling.
Yeo-Johnson scaling applies a power transformation to make the post-non-linearity features normally distributed, and thus more closely resemble the distribution of pre-non-linearity responses.
Approximate Zippering also approximates the activation function as an exponential function when re-applying it for the target model subject (step 3 of Fig.~\ref{fig:conceptual-pre-post}D).
Even approximately accounting for the activation function improves predictivity compared to linear regression, though not as much as Exact Zippering does (Fig.~\ref{fig:model_results}A).

We next compare the predictivity of Exact Zippering to the maximum achievable same-layer predictivity, estimated using a 7-layer MLP trained on 1 million response patterns driven by ImageNet-train stimuli.
The MLP does not yield substantially greater same-layer predictivity, providing some evidence that Exact Zippering is already close to the predictivity ceiling - as required for a viable IATC candidate.

\subsubsection*{\textit{Accounting for the activation function also improves area-identification specificity.}}
\label{ssec:predictivity_specificity}
A viable IATC candidate must not only align same-area responses between subjects, but also be as strict as possible, thus presumably able to separate responses from different layers.
While Exact Zippering achieves high same-layer predictivity, it is not obvious that that it should also achieve high specificity. 
For example, if Exact Zippering improves predictivity by mapping more accurately between \textit{any} pair of response profiles (including those from different layers), then we might see a decrease in specificity. 

Our primary metric for specificity is a version of the silhouette score \citep{rousseeuw1987silhouettes}.
A silhouette score close to 1 indicates that responses from different model layers (or brain areas) are well separated compared to responses from the same model layer (or brain area) (Fig.~\ref{fig:specificity}A).
For a given response profile $i$, we compute:
$$
s(i) = \frac{b(i) - a(i)}{\max(b(i), a(i))}
$$
where $a(i)$ is the mean dissimilarity between $i$ and other response profiles for the same model layer, and $b(i)$ is the mean dissimilarity between $i$ and response profiles from all other model layers. 
We take the mean score over all model subjects and layers.
As an extension of specificity, we also look at the Pearson correlation between dissimilarity scores and distances between layers in the model hierarchy (Fig.~\ref{fig:specificity}B).

Exact Zippering (and to a lesser extent, Approximate Zippering) increases specificity over linear regression and Soft Matching (Fig. \ref{fig:model_results}B).
The reason is that, by improving predictivity for the same-layer, Exact Zippering improves a key component of specificity, identification of same-layer similarity across subjects, while maintaining inter-layer separation (Fig. \ref{fig:model_results}C).

Although classical RSA (Representational Similarity Analysis) is not a predictive mapping method (instead comparing summary statistics of population responses), we can still evaluate it for specificity as a useful benchmark against which our IATC candidates can be compared, as it is widely used for neural response comparisons \citep{kriegeskorte2008representational}.
Because the RSA score uses a Pearson correlation while the mapping methods use an $R^2$ score, we also consider the squared RSA score in order to better compare to the $R^2$ scores.
We find that Exact Zippering outperforms RSA in terms of specificity (Fig.~\ref{fig:model_results}B), particularly when the RSA score is squared.

\subsubsection*{\textit{Both very strict and very flexible methods impair specificity.}}
\label{ssec:comparisons_specificity}
Soft matching, a strict method that matches individual units between populations, is not only worse for predictivity (Fig. \ref{fig:model_results}A), but also for specificity (Fig. \ref{fig:model_results}B) as evaluated using the silhouette score.
The reason is that soft matching (because of its low same-layer predictivity) has low same-layer identifiability and therefore low specificity (Fig. \ref{fig:model_results}C).
Soft matching also has lower specificity as evaluated using hierarchy correlation (Fig. \ref{fig:model_results}B). 
This is because soft matching rates adjacent layers, such as layers 2 and 3, as dissimilar even compared to more distant layers, such as layers 2 and 4 (Fig. \ref{fig:model_results}C).

At the other extreme, the flexible 7-layer MLP does not maximize specificity, because it reduces inter-layer separation, though the gain in same-layer identifiability slightly improves its specificity over ridge regression.
This result illustrates the importance of identifying the strictest set of transforms that maps accurately across subjects.
Exact Zippering and the MLP achieve similar levels of same-layer predictivity, but Exact Zippering is more constrained, leading to higher specificity.

These results illustrate the utility of the IATC, as attempting to identify it yielded a transform class that achieves high predictivity and high specificity.
In fact, our best IATC candidate approaches the Pareto frontier for predictivity and specificity (Fig.~\ref{fig:model_results}D).

\subsection*{\center{Testing IATC candidates for a mouse population}}
\label{ssec:mouse_population}

\begin{figure*}[h]
    \begin{center}
    \includegraphics[width=\linewidth]{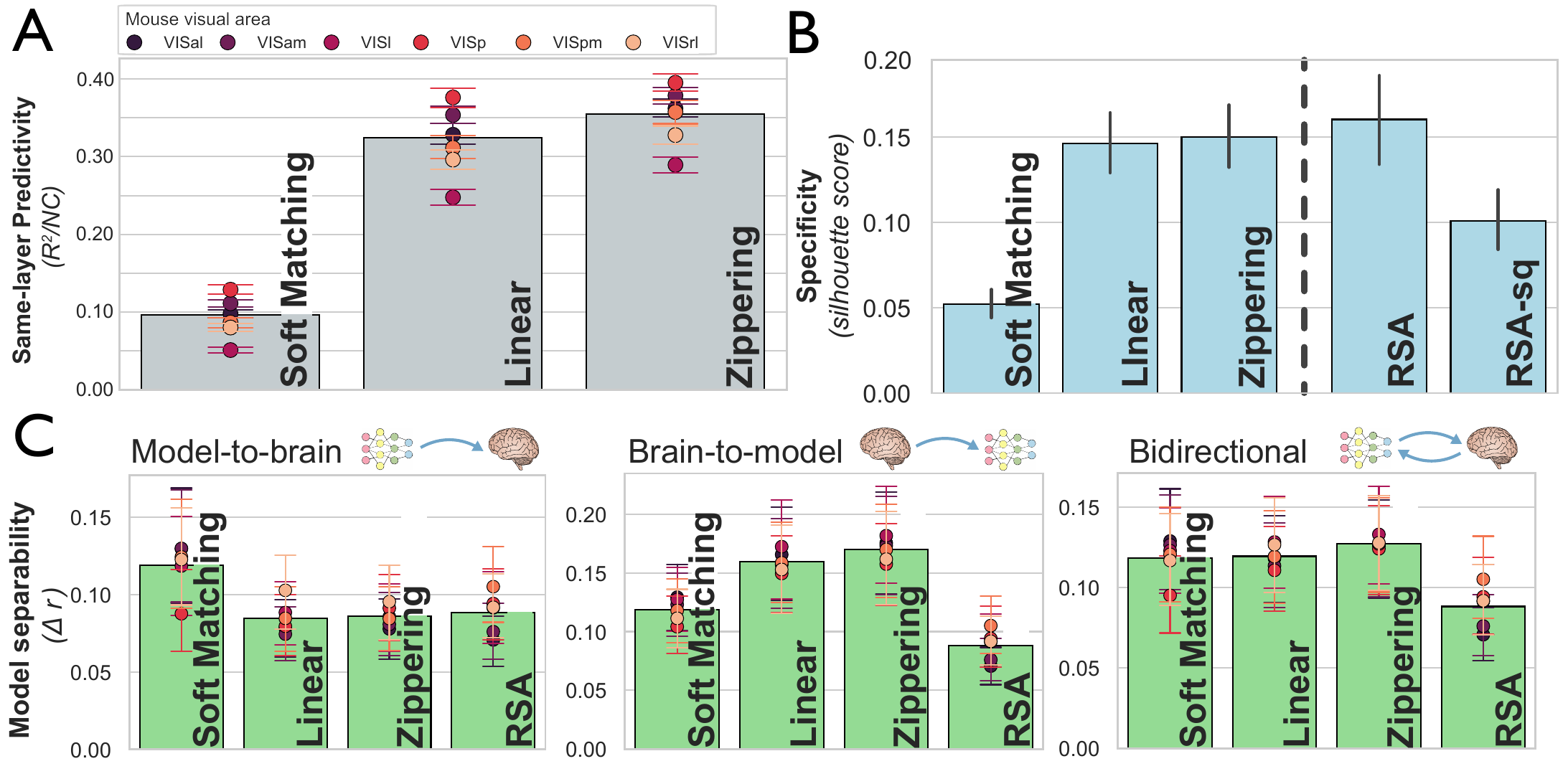}
    \end{center}
    \caption{\textbf{
    Assessing candidate transform classes on a mouse population.
    }
    (A and B) Mapping responses from pooled mouse subjects to a held-out subject in order to evaluate same-area predictivity (A) and specificity (B). 
    (C) We map five layers from four models to the mouse data: the ReLU-based AlexNet model of mouse cortex \citep{nayebi2022mouse}, our noisy softplus version of that model, ResNet, and VGG16. 
    We evaluate the mean absolute difference between models (averaged over model pairs and model layers) in terms of assessed brain similarity (using the noise-corrected Pearson correlation or RSA score).
    }
    \label{fig:mouse_results}
\end{figure*}

We now evaluate our methods on a mouse population, using Neuropixels recordings for 31 subjects in 6 brain areas averaged over 50 trials while the mice passively viewed 118 different visual stimuli.
We evaluate methods for predictivity and specificity when comparing responses between mouse subjects and also examine their ability to separate candidate models of the mouse brain.

\subsubsection*{\textit{Rank order of transform classes is largely similar between mouse and model populations.}}
The rank order of transform classes in terms of same-area predictivity provides evidence for which transform classes are better IATC candidates (despite the absolute scores being limited by relatively few response patterns for fitting the transforms, as well as a limited neuronal sample).
As in the model population, soft matching achieves the lowest same-area predictivity, with linear regression performing substantially better (Figure \ref{fig:mouse_results}A).
Moreover, our biologically motivated transform class, Zippering, that approximately accounts for the activation function further improves predictivity on the mouse data over linear regression.
The fact that a transform class that accounts for the activation function are best for same-area predictivity hints at the possibility that, just as in the simulated population, the pre-non-linearity responses of two typical mouse subject are related by a linear transform, and the relationship between post-non-linearity responses is modified by the non-linearity.

Strict methods that attempt to match individual units, such as soft matching, are worse for specificity compared to ridge regression and our biologically motivated transform classes (Figure \ref{fig:mouse_results}B).
This confirms that low predictivity can lead to low specificity by limiting same-area identifiability between subjects.
Furthermore, this confirms that our most promising candidate IATC for real neural data - Zippering - is constrained enough to separate responses from different brain areas, in addition to being flexible enough to align same-area responses between mouse subjects.

\subsubsection*{\textit{Bidirectional mapping can improve separation between brain models.}}
\label{ssec:bidirectional}
We also evaluate how well each method separates different candidate models with respect to their assessed similarity to the mouse brain responses.
We map 5 layers from four candidate models to the mouse responses: the ReLU-based AlexNet model of mouse visual cortex \citep{nayebi2022mouse}, our noisy softplus version of that model, a ResNet model trained on ImageNet categorization with 64x64 resolution stimuli, and a VGG-16 model trained on ImageNet categorization with 224x224 resolution stimuli (unlike the low resolution mouse visual system).
We apply the same noise correction procedure for predictivity scores used in \citep{nayebi2022mouse} to account for trial-to-trial variability (App.~\ref{sec:a_noise_correction_mouse}).
Model separation for a given area is evaluated as the absolute difference in assessed brain similarity between models (averaged over model pairs and model layers).

Typically, when mapping models to brains, model responses are mapped to brain responses, but the other direction of mapping is not considered.
Guided by the IATC, we map bidirectionally, just as we do when aligning two mouse brains (Figure \ref{fig:mouse_results}C).
When mapping models to brain data, stricter mappings such as soft matching  separate models more strongly, but the opposite pattern occurs when mapping brain data to models. 
When the scores for both mapping directions are averaged, the methods are all roughly comparable in terms of model separability. 
These results indicate that stricter methods like soft matching are not generally better for model separation.
Furthermore, mapping in both directions can increase separation between models compared to unidirectional mapping from model to brain.

Overall, the results in this section show that our IATC results for the model population generalize to some extent to real brain data.
In particular, we find evidence that the IATC for the mouse population is shaped by the non-linear activation function.
Moreover, we again find that a viable candidate for the IATC (Zippering) is relatively good for both prediction and specificity.
Our results also highlight the importance of the IATC's bidirectionality for model separation.

\subsection*{\center{Testing IATC candidates on a human population}}
\label{ssec:nsd}
We evaluate transform classes on a large scale human fMRI dataset, the Natural Scenes Dataset \citep{Allen2021}, and evaluate methods for predictivity and specificity when mapping between human subjects for 7 visual areas: V1, V2, V3, hV4, as well as a higher area in each of the lateral, ventral, and parietal streams.
Finally, we use IATC-guided bidirectional mapping to better separate between models of the human visual system.

\begin{figure}[t]
    \begin{center}
    \includegraphics[width=\linewidth]{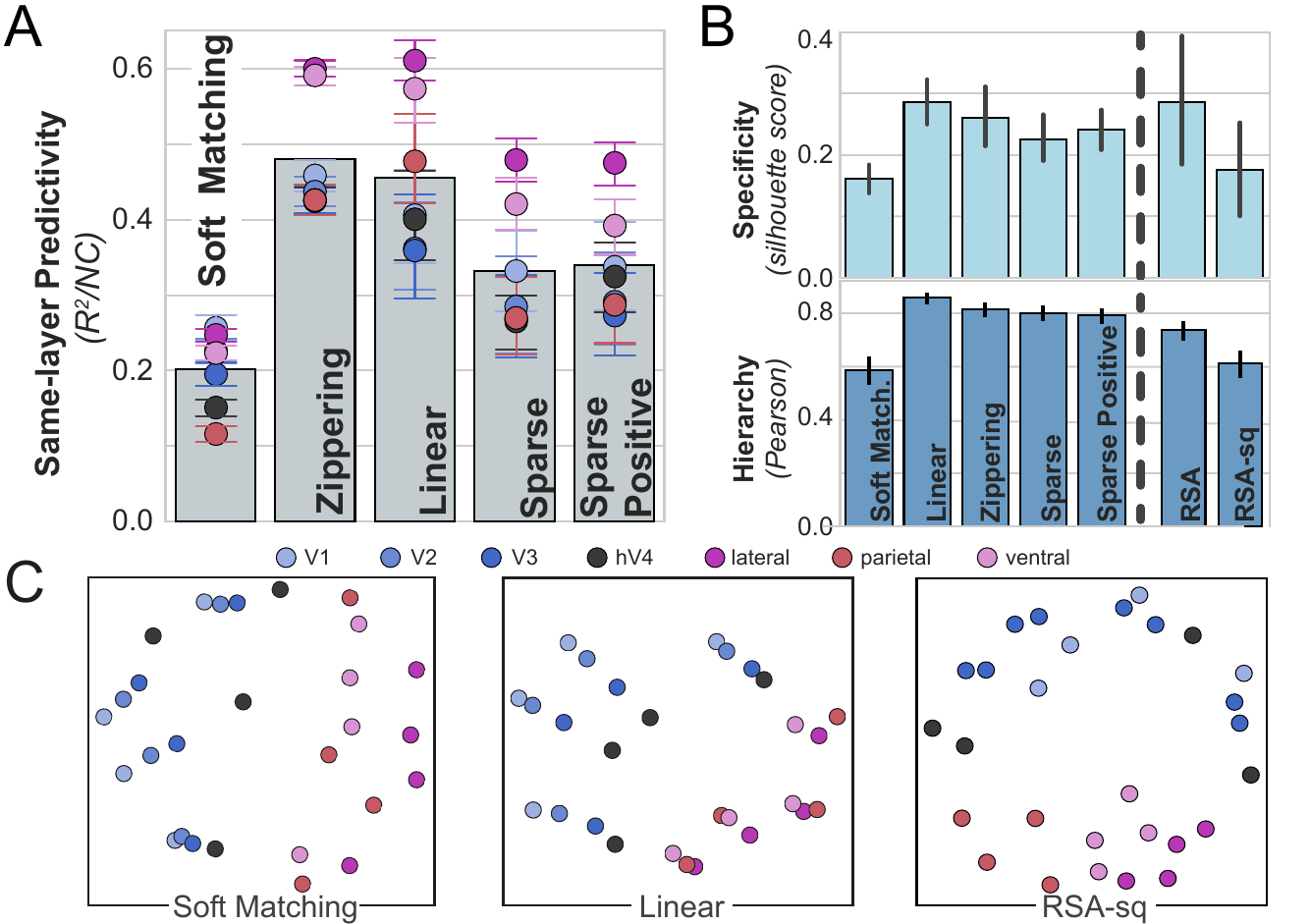}
    \end{center}
    \caption{\textbf{Assessing candidate transform classes on a human population.} 
    (A) Same-area predictivity on human subjects. 
    We map each subject to each other subject for the same brain area, using  fMRI responses to 1000 natural visual stimuli \citep{Allen2021}.
    We evaluate on 7 brain areas spanning different levels of the visual hierarchy.
    (B) Specificity and hierarchy correlation scores.
    To estimate distances in the visual hierarchy, we assign a hierarchy level of 1 to V1, 2 to V2, 3 to V3, 4 to hV4, and 5 to each of higher lateral, higher parietal and higher ventral.
    (C) MDS plots to visualize dissimilarity scores.
    }
    \label{fig:nsd}
\end{figure}

\subsubsection*{\textit{Ridge regression achieves the best intra-area cross-subject predictivity.}}
\label{ssec:nsd_coarse}

We again find that soft matching is unable to map across subjects with high predictivity.
A more flexible transform, ridge regression, is needed to map more accurately across subjects (Figure \ref{fig:nsd}A).
Although Zippering is close to ridge regression in terms of predictivity, it does not do noticeably better, perhaps because the low resolution of fMRI data obscures the effect of the non-linearity.

\subsubsection*{\textit{Ridge regression improves specificity and visual hierarchy identification.}}
\label{ssec:nsd_coarse}
Under the mean-area silhouette score, soft matching performs worse than other methods while ridge regression performs best, suggesting that soft matching has lower specificity than ridge regression (Figure \ref{fig:nsd}B).
Ridge regression scores are more correlated with distances in the visual hierarchy, suggesting that ridge regression better tracks differences across the functional hierarchy.
The improved hierarchical correlation for ridge regression compared to soft matching is apparent on an MDS plot visualizing distances between response profiles for different subjects and brain areas (Figure \ref{fig:nsd}C).

We also consider sparse regressions \citep{prince2024case}, which use a lasso penalty to encourage sparse weights (with or without a positive weights constraint).
These methods are stricter than ridge regression, but not as strict as soft matching.
We observe a loss in predictivity and specificity for sparse regressions (Fig.~\ref{fig:nsd}A,B), highlighting that even somewhat stricter-than-linear methods can impair predictivity and specificity. 
Although we cannot compare ridge regression to a very flexible control such as an MLP given the dataset size (1000 response patterns), ridge regression seems to be the best IATC candidate.


\subsubsection*{\textit{Bidirectional IATC-guided mapping improves separation of candidate brain models.}}
Recent work \citep{margalit2023unifying, finzi2022topographic} introduced a topographic model (TDANNs) of the visual system that combines functional and spatial constraints.
While the TDANN with an intermediate level of spatial loss strength $\alpha=0.25$ predicted topographical properties of visual cortex better than alternative models, a key question has been whether the TDANN's response patterns quantitatively match neuronal responses better than non-topographic models.
Here, the issue of a correct comparison method has been crucial, as unidirectional linear regression (from model to brain) failed to differentiate the TDANN from non-topographic models, while a 1-1 mapping did \citep[Fig. 6A,B]{margalit2023unifying}.
However, the very strictness of 1-1 mapping limited brain predictivity and the inter-animal noise ceiling, leaving it unclear how strong the evidence is in favor of the TDANN model.
We therefore investigated whether IATC-guided methods could distinguish between the TDANN and alternative models in terms of matching neuronal responses.

\begin{figure*}[htb]
    \begin{center}
    \includegraphics[width=\linewidth]{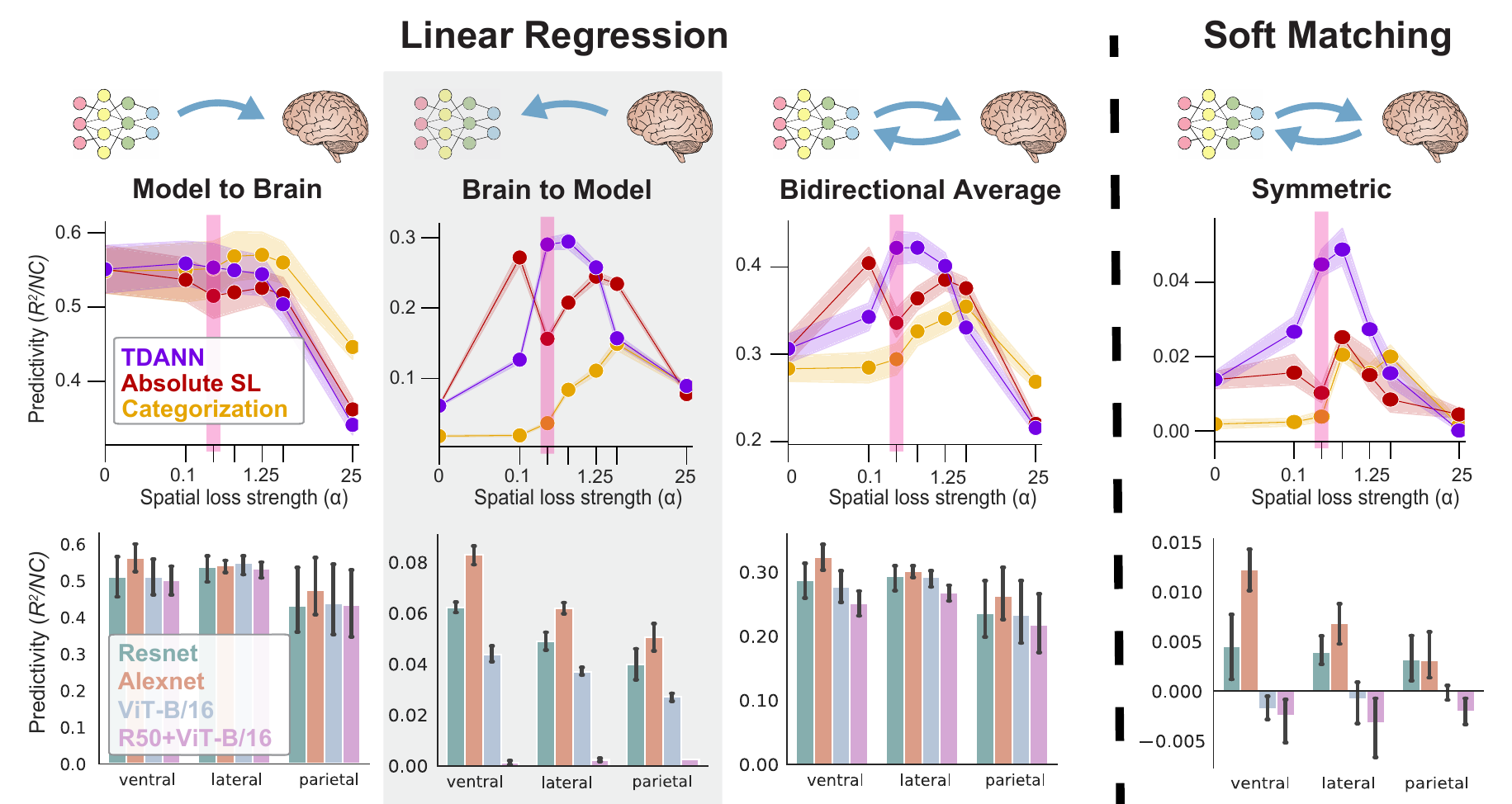}
    \end{center}
    \caption{\textbf{Using bidirectional IATC-guided mapping to improve model separation.}
    We map between models and human fMRI responses (Natural Scenes Dataset), replicating analyses from Fig. 6A,B of \cite{margalit2023unifying} and Fig.4F of \cite{khosla2023soft} that compared linear regression (model to brain direction) to stricter methods that match individual units, such as 1-1 mapping or soft matching.
    Unlike the prior analyses, we map bidirectionally between models and brains, as required by the IATC approach.
    \textit{Upper row:} Comparing topographic models \citep{margalit2023unifying} with different training objectives (TDANN, Categorization, Absolute Spatial Loss), and spatial loss strengths ($\alpha$) to higher ventral stream ROI.
    A key comparison is between the TDANN with intermediate spatial loss $\alpha = 0.25$ (highlighted with pink vertical bar) and non-topographic models ($\alpha = 0$).
    The TDANN with $\alpha = 0.25$ was found in \cite{margalit2023unifying} to best match the brain based on a one-to-one mapping and based on predicting topographic organization of the visual cortex.
    \textit{Lower row:} Comparing two CNN models (ResNet and Alexnet) and two transformer models (ViT-B/16 and R50+ViT-B/16) to higher visual areas, as in \cite{khosla2023soft}.
    }
    \label{fig:nsd_model_to_brain}
\end{figure*}

Guided by the IATC, we did ridge regression in both directions between models and the brain. 
Although linearly mapping model responses to the brain data does not separate strongly between the models, linearly mapping the brain data to the model responses separates strongly between the models (Figure \ref{fig:nsd_model_to_brain}A), identifying the TDANN with $\alpha = 0.25$ or $0.5$ as being the most brain-like, convergent with soft matching results and also with prior evidence in favor of that model \citep{margalit2023unifying}.
In fact, model separation using bidirectional ridge regression is much larger than for soft matching.
This result can be attributed to the fact that soft matching is an extremely strict method, which results in low predictivity scores for all models (Figure \ref{fig:nsd_model_to_brain}A, right-most plot).
By distinguishing more strongly between TDANN ($\alpha=0.25, 5$) and alternative models such as non-topographically constrained models ($\alpha = 0$), IATC-guided bidirectional ridge regression provides stronger evidence in favor of the TDANN.

Along similar lines, \cite{khosla2023soft} observed cases where linearly mapping models to brain responses did not separate models, but soft matching did.
Revisiting this analysis but with bidirectional mappings, we found that bidirectional ridge regression increased model separation over soft matching (Fig.~\ref{fig:nsd_model_to_brain}B), further confirming the utility of IATC-guided bidirectional mappings.
\label{ssec:nsd_fine}

%% file: sections/discussion.tex
\section{Discussion}
\label{sec:discussion}

The IATC provides a principled framework for model-brain comparison by identifying the strictest set of transforms needed to map neural responses accurately between animal subjects in a population (for the same brain area).
We find in three settings (model population, mouse population, and human population) that a working estimate of the IATC achieves both high predictivity and high specificity, two key desiderata for a model-brain comparison method.
In a simulated population of neural networks, we identified how the neuronal activation function shapes the IATC, leading to a transform class that improves predictivity and specificity relative to standard mapping classes. 
On a mouse electrophysiology dataset, we also find evidence that the IATC is constrained by the neuronal activation function, suggesting that IATC results for the model population can meaningfully generalize to real brain data.
On a human dataset, the resolution of the fMRI data does not allow us to observe an effect of the activation function on the IATC, but we still see differentiation between candidate transform classes that is consistent with our findings from simulated population and mouse data.
Moreover, we use the IATC-guided bidirectional mappings to enable better model-brain comparisons, uncovering new evidence differentiating topographic models of the visual system compared to non-topographic models.
  
Recent work has suggested that there might be a tradeoff between the goals of predictivity and specificity -- with stricter methods (such as soft matching) being better for specificity of model identification and worse at prediction, and more flexible methods (such as linear regression) showing the opposite pattern~\citep{avitanrethinking, kornblith2019similarity, ding2021grounding, conwell2022, finzi2022topographic}.
If such a tradeoff existed, it would be unclear how to map models to brains, since both of these goals are critical for model-brain comparison.
However, the intuition that stricter methods are generally better for specificity overlooks the fact that specificity requires identifying high similarity across subjects for responses of the same type, not just separating responses of different types.
Extremely strict methods fail to align same-area responses well across subjects, leading to low identifiability and thus low specificity (Fig. \ref{fig:specificity}A).
On each population, the best working estimate of the IATC improves same-area identifiability by mapping responses accurately across subjects, while still maintaining inter-area separation, leading to high specificity.
Thus, there is, in fact, no real tension between specificity and predictivity. 

A key aspect of the IATC approach is to map bidirectionally between models and brains, not just in the model to brain direction, just as when comparing two brains to each other.
Considering both directions can reveal cases where a given model contains spurious features, improving model separation.
Unlike previous works that motivated symmetry with the assumption that similarity scores must be distance metrics \citep{williams2021generalized, khosla2023soft}, under our IATC approach, symmetry naturally emerges from the mapping relationships  that exist between any two brain instances in the population (Fig.~\ref{fig:iatc_schematic}).
Our approach also differs from work that treated both directions of mapping as separate, potentially inconsistent methods \citep{soni2024conclusions} rather than as two components of a single method. 
In future work, we plan to further investigate bidirectional mappings and address potential limitations, such as how subsampling of neurons may affect the brain-to-model direction.

A limitation of our results using a simulated population is the question of whether the sources of variability we consider (different seeds for weight initialization and training data order) are anything like sources of actual brain variation. 
In future work, we will work towards a ``generative model'' that more accurately describes inter-subject variability.

\begin{wrapfigure}{r}{0.5\textwidth}
    \begin{center}
    \includegraphics[width=0.48\textwidth]{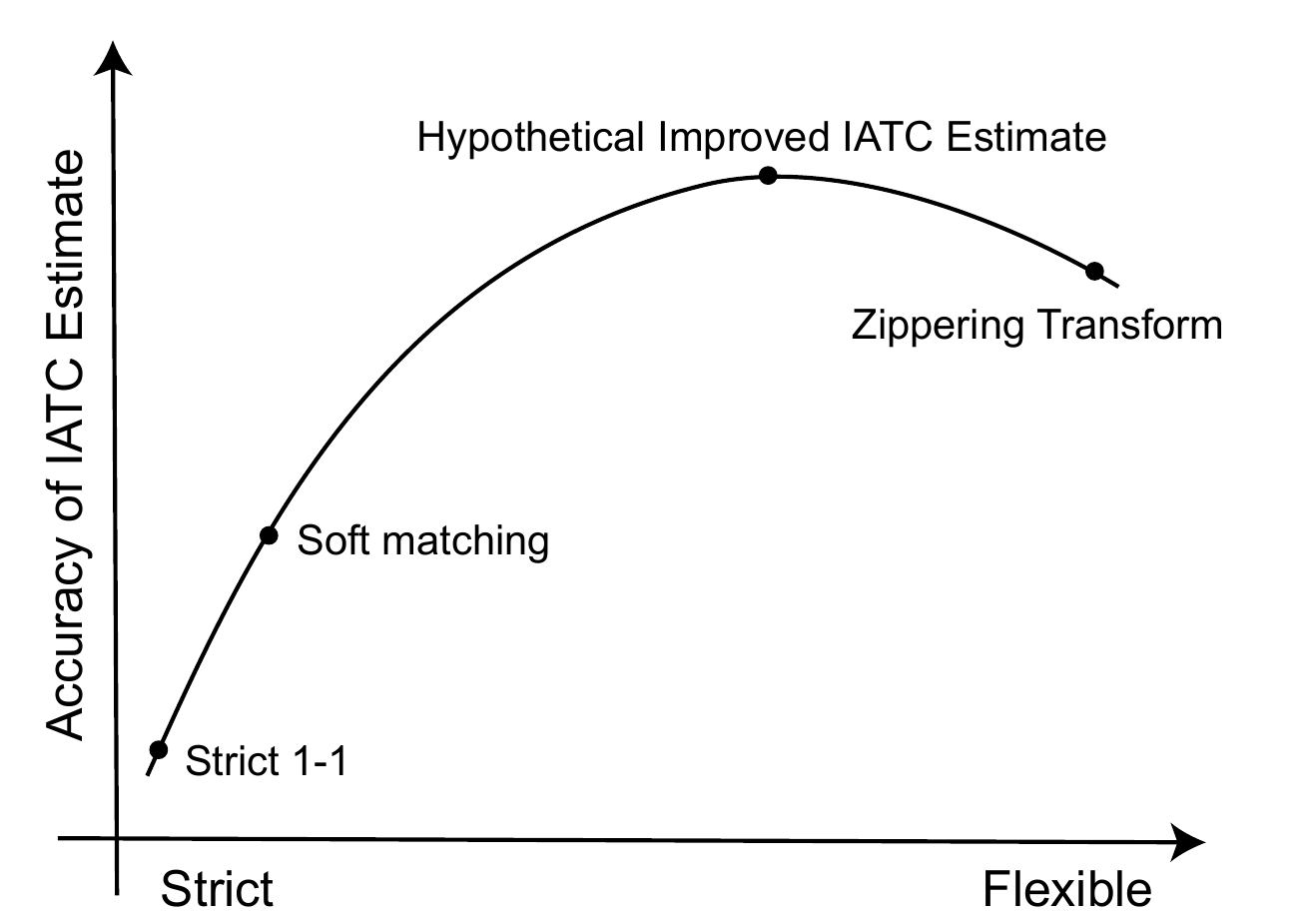}
    \end{center}
    \caption{\textbf{Taking preferred axes into account when estimating the IATC.} A hypothetical ``Preferred Axis-Zippering'' transform class could be a more accurate IATC estimate.
    }
    \label{fig:iatc_sweetspot}
\end{wrapfigure}

A second direction for improvement will be to improve IATC estimation.
Rather than evaluating a small set of transform classes as done in this paper, we hope to systematically learn the IATC in a data-driven fashion using large-scale optimization techniques.
An especially exciting possibility will be to use such data-driven estimation methods to strengthen and refine the preliminary results we show here suggesting that neural circuit mechanisms constrain the IATC.  
This will be an increasingly realistic prospect as neural datasets grow in size (with many subjects, neurons per subject, and stimuli).

Such a data-driven approach would enable us to estimate the actual IATC even in cases where the correct parcellation of the brain into distinct areas is unknown.
Because the IATC is defined to be as strict as possible (subject to the mapping accuracy requirement), the optimization process will attempt to find a transform class that can accurately match small groups of neurons across subjects (to the extent that such matchings are possible).
Thus, we expect a data-driven approach to IATC estimation to discover groups of neurons that are mappable across subjects.
These groups of neurons will likely correspond to sub-regions of brain areas that can be identified across subjects, potentially allowing for finer-grained refinements of existing parcellations, and in some cases may even correspond to distinct cell types.

Recent work has shown that neural networks, and likely the brain, have privileged axes in their representations: unit-level tuning curves that are similar between subjects, and which cannot be linearly remixed without constraint~\citep{khosla2023soft}.  
A natural synthesis of this result with our findings would be a hybrid ``Linear Subspace-Zippering'' transform class, similar to the Zippering Transform class that models the activation function, but where the linear portion of the transform is constrained by the preferred axes to specific subspaces of allowable transforms.  This hybrid would be stricter than full Linear-Zippering, but more flexible than soft matching.
We hypothesize that such a transform class would be a better estimate of the true IATC and would occupy a ``sweet spot'' on the strictness-flexibility continuum (Fig. \ref{fig:iatc_sweetspot}). 
Actually estimating these transform subspaces will likely involve novel data-driven discovery and optimization methods.

\begin{tcolorbox}[colback=white, colframe=black, boxrule=1pt, title=Key Takeaways]
\begin{enumerate}
    \item The IATC is a principled method for evaluating mechanistic models of the brain.
    \item The IATC can be implemented in practice, and it achieves both high neural predictivity and specificity of mechanism identification, as required for accurate model-brain comparisons.
    \item Estimating the IATC correctly requires taking into account details of the neural mechanism, such as the non-linear activation function present in the visual cortical hierarchy.
\end{enumerate}
\end{tcolorbox}

%% file: sections/suppl.tex
\vspace{1em}
{\centering\large\textbf{--- Supplementary Materials ---}\par}
\vspace{1em}


\section{Central hypercolumn selection}
\label{sec:a_central}
In order to map between units with similar functional roles (at least for the same model layer), we do our model-model fits using only the central hyper-column of units in each layer (i.e. the units whose receptive field is directly at the middle of the input image).
Indeed, even when constraining the mapping to use only the central hyper-column, we are able to identify high similarity across model instances for the same layer, at least when assessing pre-non-linearity responses using a linear transform.

\section{Soft matching as a transform class}
\label{sec:a_soft_matching}
While \citet{khosla2023soft} do not explicitly formulate the soft matching score as a predictive mapping, it can be formulated as one.
Computing the soft matching score involves maximizing:
$$
\Sigma_{i, j} \mathbf{T}_{ij} \textbf{C}_{ij}
$$
where $\mathbf{T}$ is the transport matrix, subject to the constraints that the columns of the matrix sum to $1/N_Y$, while the rows sum to $1/N_X$, and $\mathbf{C}$ is the matrix of Pearson correlations between each source neuron and each target neuron.
The transport matrix can be interpreted as a joint probability distribution over source neurons and target neurons (where the marginal distributions are uniform discrete).
Thus, the soft matching score is the  \textit{expected} correlation between source and target neurons, according to joint probabilities encoded by the optimal transport matrix.

Since maximizing the above objective requires identifying source neurons that are highly correlated with each target neuron, we can use the source neurons to predict the value of each target neuron, weighted by the probabilities in the optimal transport matrix. 
First, for each source neuron $X_i$ and target neuron $Y_j$, we can predict $Y_j$'s responses across a set of stimuli (symbolized as the vector $\mathbf{Y}_j$) based on $X_i$'s responses to those stimuli (symbolized as $\mathbf{X}_j$) as:
$$
\mathbf{\hat{Y}}_j = \frac{\sigma(\mathbf{Y}_j)}{\sigma(\mathbf{X}_i)}[\mathbf{X}_i - \bar{\mathbf{X}_i}] \mathbf{C}_{ij} + \bar{\mathbf{Y}_j}
$$
This is essentially using the correlation $\mathbf{C}_{ij}$ to do ordinary least squares between $X_i$'s responses and $Y_j$'s responses.

For a single target neuron $Y_j$, we compute the expected value of these correlation-based predictions across source neurons, if we sampled source neurons according to the conditional probability distribution $P(X = X_i | Y = Y_j)$.
Since $\mathbf{T}_{ij} = P(X_i, Y_j)$ and $P(Y_j) = 1/N_Y$, it follows that $P(X = X_i | Y = Y_j) = N_Y \mathbf{T}_{ij}$ .
Using these conditional probabilities, the overall prediction $\mathbf{\hat{Y}}_j$ then becomes: 
$$
\mathbf{\hat{Y}}_j = N_Y \sigma(\mathbf{Y}_j)\Sigma_i \frac{\mathbf{X}_i - \bar{\mathbf{X}_i}}{\sigma(\mathbf{X}_i)} \mathbf{T}_{ij} \mathbf{C}_{ij} + \bar{\mathbf{Y}_j}
$$


\section{Motivating the softplus activation function with a simple model of a noisy spiking process}

\begin{figure}[h!]
    \begin{center}
    \includegraphics[width=\linewidth]{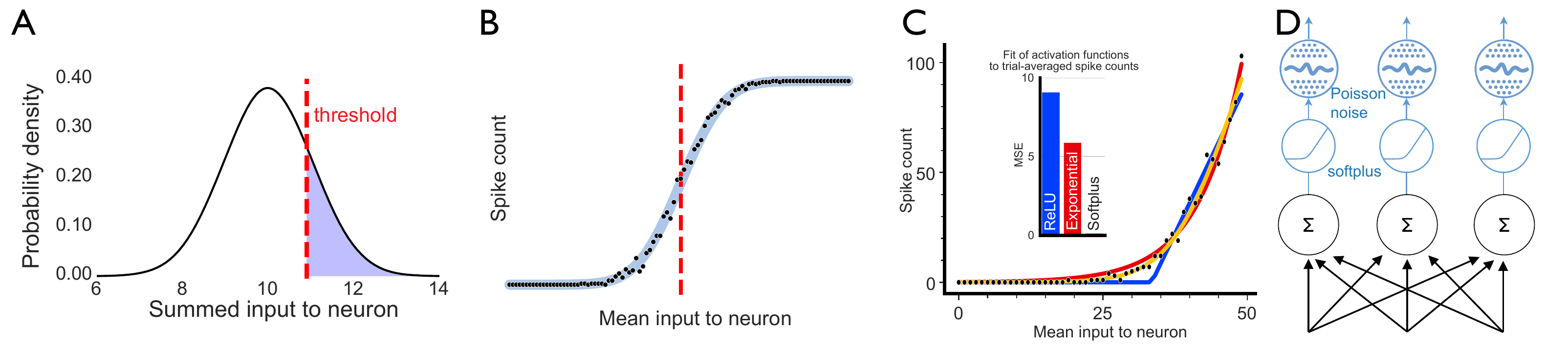}
    \end{center}
    \caption{
\textbf{    A more biologically consistent activation function.}
    (A) Biological activation functions are the result of a noisy spiking process. Because summed inputs to neurons are noisy, the firing probability is positive even when the mean input is sub-threshold. 
    Here, the probability of spiking is represented as the size of the blue region.
    (B) The resulting activation function, unlike ReLU, is strictly positive and increasing. 
    Dots represent simulated spike counts, which are Poisson-distributed in the limit of very small firing rates.
    (C) Fitting different activations to simulated spike counts, allowing for scaling and translation. 
    Softplus fits spike counts the best in the sub-threshold regime. 
    The exponential activation also function performs somewhat better than ReLU. 
    Intuitively, the reason ReLU does not fit as well is that it has a hinge that prevents it from capturing the smooth increase in firing rate. 
    Spike counts are plotted for a single trial.
    (D) We replaced each ReLU non-linearity in the models with a softplus non-linearity and a Poisson-like noise sampler.
    }
    \label{fig:biologically_consistent_activation_function}
\end{figure}

\label{sec:a_noisy_spiking}
Our simulation of spike counts is based on the following highly simplified model.
We assume that a neuron receives total input $X \sim \mathcal{N}(\mu, \sigma^2)$ and that during a single time interval equal to the neuron's refractory period (which we assume to be about 1 ms), the neuron either fires once or not at all, depending on whether $X > T$, where $T$ is a fixed threshold (Fig.~\ref{fig:biologically_consistent_activation_function}A).
We count the number of spikes over a 100 ms time range, and average over 100 trials.

Under this model, the total mean (over trials) spike count $S_t(\mu)$ over a time period $t$ (expressed as a function of the mean total input to the neuron $\mu$) is equal to $t/R * \Phi(\mu - T, \sigma^2)$, where $\Phi$ is the Gaussian CDF.
This means that the activation function should have a sigmoid shape, which saturates at sufficiently high mean inputs (Fig.~\ref{fig:biologically_consistent_activation_function}B).
However, many cortical neurons are thought to fire in the fluctuation driven, unsaturated regime \citep{van1996chaos}.
We therefore focus on unsaturating functions like softplus and fit these functions to spike counts that we simulated in the unsaturated regime (Fig.~\ref{fig:biologically_consistent_activation_function}C).
The softplus function is defined as:
$$
\text{softplus}(x) = \ln(1 + e^x)
$$

\section{Noisy Softplus AlexNet models}
\label{sec:a_noisy_models}
To obtain our noisy softplus variant of the AlexNet mouse model, every ReLU sub-layer in the AlexNet models is exchanged for a Softplus sub-layer followed by a Poisson-like noise block whose mean is the output of the Softplus sub-layer. 
PyTorch enables noisy models to be trained using a reparameterization trick, but only for certain probability distributions (not for the Poisson distribution).
We use the Gamma distribution as a stand-in for Poisson, choosing shape parameter $k = \lambda$, where $\lambda$ is the Poisson parameter (which is chosen to be the output of the Softplus sub-layer), and scale parameter $\theta = 1$.
This allows us to replicate two statistical properties of Poisson variables: non-negative samples and variance-mean ratio of 1, both of which are important for using the Zippering Transform or its approximate variant (which both use a Poisson GLM) to predict the responses.
To avoid numerical difficulties for small values of $k = \lambda$, we scale the softplus outputs by 100 before sampling from the Gamma distribution.
We then train the noisy softplus models so that their instance recognition training score (as well as validation score on ImageNet categorization) are equal to those of the ReLU-based AlexNet models. 

\section{Inverting the softplus function in the Zippering Transform}
\label{sec:softplus_inverse}
In order to invert the softplus non-linearity as the first step of the Zippering Transform, we apply the inverse of the softplus function to the softplus model responses in a given layer, averaged over 50 trials.
Because the softplus outputs at every model layer are scaled by 100 before taking Poisson-like samples from the Gamma distribution (App.~\ref{sec:a_noisy_models}), we un-scale the trial-averaged responses before applying the softplus inverse.
The inverse of the softplus function is well-defined (because softplus is strictly increasing) and has the following formula:
$$
\text{softplus}^{-1}(y) = \ln(e^y - 1)
$$
In practice, to avoid numerical difficulties for very small values of $y$, we do not apply this formula directly and instead use a more numerically stable implementation of the softplus inverse adapted from the TensorFlow library \citep{tensorflow2015-whitepaper}.

\section{Yeo-Johnson scaling in Approximate Zippering}
\label{sec:a_yeo}
\begin{figure}[h]
    \begin{center}
    \includegraphics[width=0.5\linewidth]{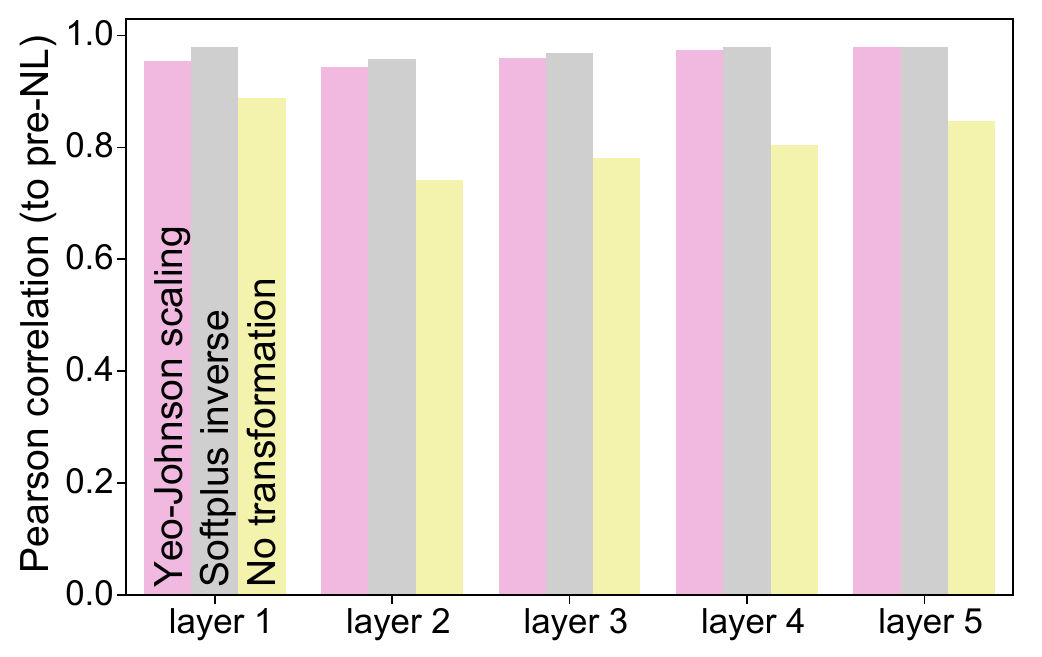}
    \end{center}
    \caption{\textbf{Correlation between post-non-linearity responses and pre-non-linearity responses after transforming the post-non-linearity responses in different ways (responses are for the noisy softplus models, averaged over 50 trials).}
    We focus on correlation here because Yeo-Johnson scaling does not improve the $R^2$ score with respect to pre-non-linearity features (i.e. it does not directly match them), which makes sense as it is merely unskewing the distribution of post-NL features, which are already rather correlated with pre-NL features.
    Nevertheless, increased correlation implies that the pre-NL features can be more easily matched after linear re-weighting, as is done in Approximate Zippering.
    }
    \label{fig:yj_scaling}
\end{figure}

When the activation function is known exactly and its inverse is well-defined (as in the case of the Softplus-based model), we can directly invert the activation function to recover the pre-non-linearity responses.
However, when mapping animals to animals (or, if enough neurons are measured, animals to models), we cannot easily invert the activation function if we do not know its exact form for a given neuron.
Yeo-Johnson scaling uses a power transformation to make the features closer to normally distributed over the stimuli.
We expect this transformation to make the post-non-linearity features more correlated with pre-non-linearity responses because the non-linear activation function skews the distribution of the pre-non-linearity responses (which are roughly normally distributed over the stimuli).
Indeed, we find that Yeo-Johnson scaling noticeably increases the Pearson correlation (Fig. \ref{fig:yj_scaling}) with the pre-non-linearity responses for the noisy softplus models, almost as much as if you had directly applied the inverse of the softplus activation function to the post-non-linearity responses.
We hypothesize that Yeo-Johnson scaling has a similar effect in the case of animal firing rates.

We implement Yeo-Johnson scaling with the PowerTransformer class in sklearn \citep{scikit-learn}.
The power transform fits one parameter.
To implement Yeo-Johnson scaling as the first step of Approximate Zippering, we put the PowerTransformer object followed by a GLM object into an sklearn Pipeline, so that the power parameter is only fit on the training data, not on test data.

\section{Implementation details of GLMs}
\label{sec:a_glm}
The GLM object is created using the \textit{glum} package \citep{thompson2025}.
Each GLM specifies the inverse link function that relates the linear prediction to the response variable (such as ReLU, exponential or softplus), and the assumed noise structure in the response variable (Poisson noise in the case of Approximate Zippering or the Zippering Transform). 
The weights of the specified GLM are then optimized through Iterative Reweighted Least Squares.

The inverse link function in the Zippering Transform  involves a scaling parameter $c$:
$$
\hat{y} = c*\text{softplus}(\theta^T x)
$$
where $\hat{y}$ is the output of the inverse link function (i.e. the predicted values for the target responses), $x$ is the vector of predictors (trial averaged responses of the source model after inverting the activation function) and $\theta$ is the fitted linear weights.
When predicting noisy softplus model responses, we set $c = 100$, the same softplus output scaling we used when training the models themselves.

\section{Noise correction when comparing models to mouse data}
\label{sec:a_noise_correction_mouse}
When mapping between model responses and trial-averaged mouse responses (Fig.~\ref{fig:mouse_results}C), it is important to account for trial-to-trial variability.
Here we briefly describe the noise correction procedure that is used to obtain more accurate predictivity scores.
The full derivation of this procedure is found in \cite{nayebi2022mouse}.
The goal of the procedure is to accurately estimate the following quantity:
$$\operatorname{Corr} (\mathcal{M}(r_{\text{train}}; t^B_{\text{train}})_{\text{test}}, t^B_{\text{test}})$$ 
where $\mathcal{M}$ is a given mapping method (such as ridge regression), $r_{\text{train}}$ is the model responses in a given layer (which, except for the noisy softplus models, are deterministic) over the training stimuli, $t^B_{\text{train}}$ is the \textit{true} trial-averaged (averaged over the ideal limit of infinitely many trials) responses of a particular subject and brain area $B$ over training stimuli, $\mathcal{M}(r_{\text{train}}; t^B_{\text{train}})_{\text{test}}$ are the test predictions under the mapping method of the target animal's responses over the test stimuli using the model responses as predictors, and $t^B_{\text{test}}$ are the actual ground-truth responses of the target animal over test stimuli. 
This quantity cannot be directly computed because we do not have infinitely many trials per stimulus, and instead must estimate it based on finitely many trials (50 trials per stimulus in the case of the Allen Institute mouse data).

To perform the noise correction, we use bootstrapping.
For each bootstrapped sample, we separate the $N=50$ trials into two split halves of 25 trials each (indexed in the notation given below by 1 and 2) and take the trial-averaged response for each stimulus for each split half of those trials.
Then the noise-corrected predictivity (in terms of Pearson correlation) is computed as:
$$
\operatorname{median} \left\langle \frac{\operatorname{Corr} \left( \mathcal{M}(r_{\text{train}}^\ell ; s_{1,\text{train}}^B)_{\text{test}}, s_{2,\text{test}}^B \right)}{\sqrt{\widetilde{\operatorname{Corr}} \left( \mathcal{M}(r_{\text{train}}^\ell ; s_{1,\text{train}}^B)_{\text{test}}, \mathcal{M}(r_{\text{train}}^\ell ; s_{2,\text{train}}^B)_{\text{test}} \right) \times \widetilde{\operatorname{Corr}} \left( s_{1,\text{test}}^B, s_{2,\text{test}}^B \right)}} \right\rangle
$$
where the median is computed over the target animal's neurons, and the $\langle ... \rangle$ represents an average over all bootstrap samples. The $\widetilde{\operatorname{Corr}}$ represents a Spearman-Brown corrected Pearson correlation rather than a raw Pearson correlation.
$s^B_{\text{i, train/test}}$ represents the trial-averaged responses of the subject for split-half $i$ (which is either 1 or 2) over the train stimuli or test stimuli (unlike $t$ which was the ideal trial-average over infinitely many trials).

In most cases, we use 100 bootstrapping samples, and 10 train-test splits. However, in some cases, we use fewer bootstrapping samples and train-test splits because of computation time constraints. In particular, whenever we map from VGG-16 to mouse responses, or whenever we use Linear Nonlinear, we use 16 bootstrapping samples and 1 train-test split.

\section{Noise correction when comparing models to human fMRI data}
\label{sec:a_noise_correction_human}
The bootstrapping approach to noise correction described in App.~\ref{sec:a_noise_correction_mouse} is not possible in the case of the human fMRI data, where there are only 3 trials per stimulus, not 50 trials.
We instead use the method of noise correction recommended in \cite{Allen2021}.
The idea is to simply divide the raw $R^2$ predictivity (with respect to a given target voxel) by the noise ceiling of that target voxel, which is computed as:
$$
NC = \frac{\text{ncsnr}^2}{\text{ncsnr}^2 + \frac{1}{n}}
$$
where ncsnr stands for ``noise ceiling signal-to-noise ratio" and is provided for each voxel with the Natural Scenes Dataset, and $n$ is the number of trials (3).
This accounts for trial-to-trial variability when mapping model units to noisy voxel responses. 
When mapping in the other direction, from noisy voxels to model units, we set $NC = 1$, since the models of the human visual system we consider are all deterministic.
It is worth noting that this procedure can only account for noise in the target voxel, but cannot account for noise in the source predictors (which is certainly an issue when either mapping brain-to-brain or brain-to-model).
Developing statistical methods to correct for source noise is a major open challenge for future research.